\documentclass[letterpaper, conference]{IEEEtran} %

\usepackage{pifont}
\newcommand{\cmark}{\ding{51}}
\newcommand{\xmark}{\ding{55}}

\usepackage{algorithm}
\usepackage{algpseudocode}
\usepackage{amsmath,amssymb,amsfonts,amsthm}
\usepackage{epsfig}
\usepackage{graphicx}
\usepackage{textcomp}
\usepackage{xcolor}
\usepackage{booktabs}
\usepackage{cite}
\usepackage{url}
\usepackage{multirow}
\usepackage{subcaption}
\usepackage{mathtools}
\usepackage{bbm}
\usepackage{nicefrac}

\newcommand{\beq}{\begin{equation}}
\newcommand{\eeq}{\end{equation}}

\begin{document}

\title{\Huge ChronoSC: Task-Oriented Semantic Communication via Temporal-to-Color Encoding\vspace{-5pt}}

\author{
    \IEEEauthorblockN{
        Phuc H. Nguyen$^1$, 
        Trung T. Nguyen$^1$, 
        Quy N. Duong$^1$, 
        and Van-Dinh Nguyen$^{2}$}
    \IEEEauthorblockA{\it $^1$Smart Green Transformation Center (GREEN-X), VinUniversity, Vietnam. \\
     $^2$School of Computer Science and Statistics, Trinity College Dublin,  Ireland\\
    E-mails: \{23phuc.nh, 22trung.nt, 23quy.dn\}@vinuni.edu.vn, dinh.nguyen@tcd.ie
    \vspace{-5pt}}
    \footnote{}

}

\maketitle

\begin{abstract}
Semantic communication (SC) aims to reduce transmission overhead by conveying task-relevant information rather than raw data. However, existing SC approaches for video largely focus on pixel-level reconstruction or rely on complex spatiotemporal pipelines, leading to excessive bandwidth usage and latency that are unsuitable for low-resource deployments. In this paper, we propose ChronoSC, a task-oriented semantic communication framework for Video Question Answering (VideoQA). ChronoSC introduces Chrono-Color Stacking, a lightweight and lossless projection scheme that encodes temporal video dynamics into a single static image, enabling extreme temporal compression before transmission. This compact semantic representation is transmitted using a lightweight Deep Joint Source-Channel Coding (DeepJSCC) transceiver and explicitly reconstructed at the receiver. Unlike latent-space methods, explicit visual reconstruction enables the direct reuse of pre-trained vision-language models; specifically, a pre-trained BLIP model is employed to infer answers from noisy, reconstructed chrono-images. Experiments on the CLEVRER dataset show that ChronoSC achieves up to 192 times bandwidth reduction compared to raw video transmission while maintaining high VideoQA accuracy.
\end{abstract}


\section{Introduction}Video data currently dominates global internet traffic, driven by the rapid proliferation of Internet of Things (IoT) devices, unmanned aerial vehicles (UAVs), and autonomous systems. In many emerging applications, such as intelligent traffic monitoring, surveillance, and hazard detection, the primary objective is no longer to deliver visually appealing video streams for human consumption. Instead, the goal is for machines to extract actionable intelligence directly from visual data, such as object attributes, motion patterns, or causal events \cite{SenuraTMC}.

Despite this shift toward machine-centric video analytics, transmitting video from edge devices to a central server remains fundamentally challenging. In particular, two key constraints limit the practicality of large-scale deployments. First, transmitting full video sequences requires high data rates that often exceed the capacity of edge and wireless networks. Second, conventional video coding standards (e.g., H.264/AVC and H.265/HEVC) suffer from the well-known “cliff effect” \cite{cliff}, where even a modest degradation in the signal-to-noise ratio (SNR) can cause catastrophic decoding failure, rendering the received video unusable for downstream analysis.

These challenges expose a fundamental mismatch between traditional video communication systems, designed primarily for human viewing, and the requirements of machine-oriented inference tasks. To address this gap, semantic communication (SC) has emerged as a promising paradigm that aims to transmit only task-relevant semantic information rather than raw pixel data \cite{semcom_survey}. While recent works, such as VideoQA-SC \cite{videoqasc}, have extended SemCom to video-based reasoning tasks, they typically rely on complex spatiotemporal processing pipelines built on 3D convolutional networks, object detectors, or graph-based reasoning modules \cite{wu2022slotformer},\cite{le2020hierarchical}. Such architectures introduce substantial computational overhead and latency, making them impractical for resource-constrained edge devices.

Building on this insight, we propose ChronoSC, an ultra-low-complexity video semantic communication framework for VideoQA. ChronoSC introduces Chrono-Color Stacking, a parameter-free projection method that deliberately leverages fundamental arithmetic operations (e.g., background subtraction and progressive hue shifting) to achieve near-zero computational overhead. By explicitly encoding temporal dynamics into spatial color trails within a single RGB image, we create a compact semantic representation that bypasses the need for intensive feature extraction at the edge. This representation is transmitted using the proposed Motion-Aware Swin Transceiver (MAST), a deep joint source-channel coding architecture that prioritizes dynamic regions for robust transmission. Unlike latent-space methods, MAST explicitly reconstructs a pixel-domain semantic image at the receiver, enabling direct reuse of pre-trained vision-language models, such as BLIP \cite{blip}, to infer answers from noisy reconstructions without retraining the transmitter.

The main contributions of this paper are summarized as follows:
\begin{itemize}
    \item We propose ChronoSC, an ultra-lightweight projection scheme that compresses temporal video dynamics into a single static image. Unlike existing methods relying on heavy deep learning backbones, our parameter-free design eliminates the need for spatiotemporal feature extraction, resulting in near-zero computational overhead (measured in FLOPs) suitable for extreme edge deployments.
    \item We develop a modular semantic interface via the proposed MAST transceiver. By introducing a motion-aware gating mechanism, MAST explicitly prioritizes dynamic regions for robust transmission. Furthermore, its pixel-domain reconstruction explicitly decouples the communication module from downstream reasoning models, enabling the "plug-and-play" reuse of Foundation Models (e.g., BLIP).
    \item Experimental results on the CLEVRER dataset \cite{clevrer} demonstrate that ChronoSC achieves a massive $192\times$ bandwidth reduction compared to raw video transmission. Remarkably, under extreme channel noise ($\text{SNR} = 0$ dB), ChronoSC maintains a high VQA accuracy of $76.2\%$, significantly outperforming conventional digital codecs (H.264 + LDPC) which suffer from the cliff effect and drop to $35.2\%$.
\end{itemize}

\section{Related Work}

Traditional communication systems follow a strict separation between source coding for compression and channel coding for error protection. Deep joint source-channel coding (DeepJSCC) replaces this modular design with neural networks trained end-to-end, enabling more efficient and robust transmission over wireless channels. Bourtsoulatze \textit{et al.} \cite{bourtsoulatze} first demonstrated that CNN-based JSCC can effectively mitigate the cliff effect in wireless image transmission. Building on this idea, SwinJSCC \cite{swinjscc} incorporates Swin Transformers to better capture global contextual information, further improving reconstruction quality and robustness. These advances highlight the effectiveness of transformer-based JSCC for image transmission, which we adopt and extend in this work.

Unlike conventional communication systems that aim for faithful signal reconstruction, task-oriented semantic communication focuses on maximizing the performance of a downstream task, such as classification or Visual Question Answering (VQA). MU-DeepSC \cite{mudeepsc} jointly encodes image and text semantics for VQA but is limited to static images and does not address temporal dynamics in video. VideoQA-SC \cite{videoqasc} extends semantic communication to video but relies on computationally intensive spatiotemporal encoders that combine object detection, temporal modeling, and graph neural networks. In contrast, our approach shifts complexity away from the network architecture toward the data representation itself. By encoding temporal dynamics into a compact 2D Chrono-Color image, we enable the use of lightweight image-based models rather than heavy video processing pipelines.

Representing motion within a single image has been explored in computational photography and computer vision. Yosef \textit{et al.} \cite{yosef} used coded exposure to enable video reconstruction from a single motion-blurred image, while VDM-MD \cite{vdmmd} employed diffusion models to resolve temporal ambiguity from blurred inputs. These methods, however, are designed for high-fidelity video reconstruction for human viewing. In contrast, our work is fundamentally task-oriented: we do not seek to reconstruct the original video, but instead treat the stacked image as a semantic symbol that directly encodes motion information for machine interpretation.

\section{System Model and Problem Formulation}
\begin{figure*}[!t]
    \centering
    \includegraphics[width=\linewidth]{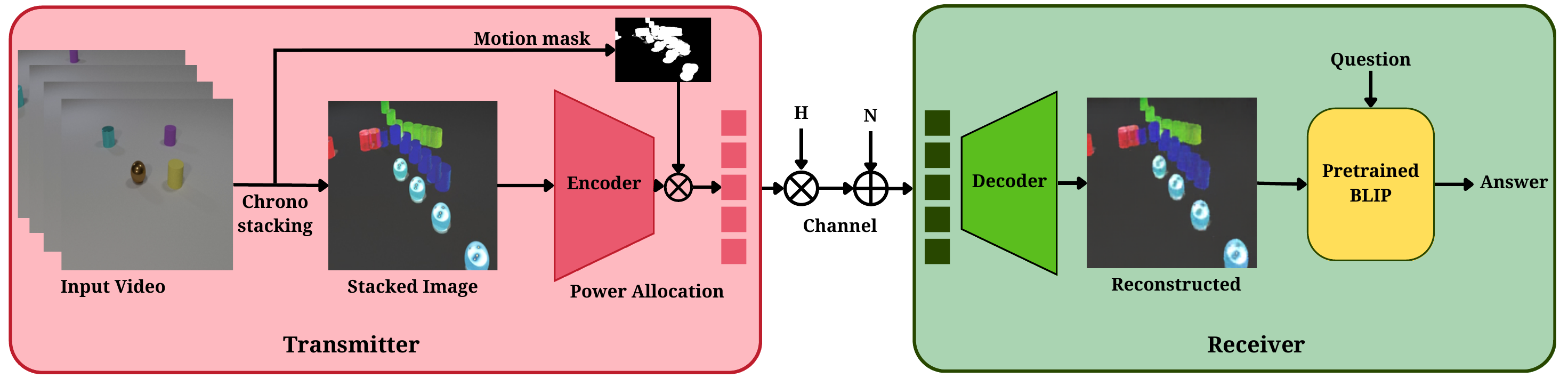}
    \caption{End-to-end SC framework for ChronoSC, where the input video is compressed into a single stacked semantic image and transmitted for VideoQA inference.}
    \label{fig:System_Diagram}
\end{figure*}

Fig. \ref{fig:System_Diagram} illustrates the proposed end-to-end SC framework for ChronoSC, designed for Video Question Answering (VQA) at the network edge. The system consists of a semantic transmitter, a wireless channel, and a semantic receiver, and aims to efficiently transmit task-relevant visual semantics under bandwidth and channel constraints.

\noindent\textbf{Transmitter}: The input is a video sequence $\mathbf{V} \in \mathbb{R}^{T \times 3 \times H \times W}$, where $T$ denotes the number of frames, and $H$ and $W$ represent the spatial height and width, respectively. Instead of transmitting the full video, the transmitter first extracts a compact semantic representation and then encodes it:
\begin{enumerate}
    \item \textit{Semantic projection:} A projection function $\mathcal{P}(\cdot)$ compresses the video into a single static semantic representation $\mathbf{I}_{\text{sem}}$ and simultaneously extracts a binary motion mask $\mathbf{M} \in \{0, 1\}^{H \times W}$:
    $ (\mathbf{I}_{\text{sem}}, \mathbf{M}) = \mathcal{P}(\mathbf{V}).$
    The semantic image encodes temporal dynamics, while the motion mask identifies spatial regions associated with motion.
    
    \item \textit{Motion-aware encoding:} The semantic image $\mathbf{I}_{\text{sem}}$ is encoded into a sequence of $k$ complex channel symbols $\mathbf{Z} \in \mathbb{C}^k$ using the MAST encoder $f_{\theta}(\cdot)$, which leverages the motion mask to prioritize dynamic regions:
    $\mathbf{Z} = f_{\theta}(\mathbf{I}_{\text{sem}}, \mathbf{M}).$
    The transmitted symbols are normalized to satisfy the average power constraint: $\frac{1}{k} \mathbb{E}[\|\mathbf{Z}\|^2] \le 1$.
\end{enumerate}

\noindent\textbf{The channel model}: The modulated symbols $\mathbf{Z}$ are transmitted over a wireless fading channel. The received signal $\hat{\mathbf{Z}} \in \mathbb{C}^{k}$  is modeled as: $ \hat{\mathbf{Z}} = \boldsymbol{h} \cdot \mathbf{Z} + \mathbf{n}$,
where $\boldsymbol{h}$ represents the channel coefficient and $\mathbf{n} \sim \mathcal{CN}(0, \sigma^2 \mathbf{I})$ denotes the Additive White Gaussian Noise (AWGN) with noise power $\sigma^2$. 

\noindent\textbf{Receiver}:
The receiver infers the correct answer $A$ directly from the noisy received symbols $\hat{\mathbf{Z}}$ in two sequential stages:
\begin{enumerate}
    \item \textit{Visual reconstruction:} The MAST decoder $g_{\phi}(\cdot)$ first reconstructs a pixel-domain semantic image:
    $\hat{\mathbf{I}}_{\text{sem}} = g_{\phi}(\hat{\mathbf{Z}}).$
    
    \item \textit{Semantic inference:} The reconstructed image is passed to a VQA model  $\mathcal{M}_{\psi}(\cdot)$ (e.g., BLIP) together with the question $Q$, to predict the answer:
$ \hat{A} = \mathcal{M}_{\psi}(\hat{\mathbf{I}}_{\text{sem}}, Q).$
\end{enumerate}

\noindent\textbf{The problem formulation}:
ChronoSC seeks to maximize VQA accuracy while minimizing bandwidth usage. We define the Bandwidth Compression Ratio (BCR) as
\begin{equation}
    \text{BCR} = \frac{k}{T \times H \times W \times 3}.
\end{equation}

To enable modularity and reuse of pre-trained vision-language models, training is decoupled into two sub-problems:
\begin{enumerate}
    \item \textbf{Reconstruction fidelity (MAST):} The MAST encoder–decoder pair is trained to minimize a motion-weighted reconstruction distortion between the original and reconstructed semantic images:
\begin{equation}
    \min_{\theta, \phi} \mathbb{E}\left[ \|\mathbf{I}_{\text{sem}} - \hat{\mathbf{I}}_{\text{sem}}\|^2 \odot (\mathbf{1} + \alpha \mathbf{M}) \right], \, \texttt{s.t.} \, \text{BCR} \le \gamma
\end{equation}
where $\alpha$ weights the motion importance and $\gamma$ is the bandwidth constraint.
    \item \textbf{Task adaptation (VLM):} Independently, the VQA model is trained on clean semantic images to minimize the cross-entropy loss:
\begin{equation}
    \min_{\psi} \mathcal{L}_{\text{CE}}(\mathcal{M}_{\psi}(\mathbf{I}_{\text{\text{sem}}}, Q), A_{\text{gt}})
\end{equation}
\end{enumerate}
where $A_{\text{gt}}$ is the ground-truth answer.

By applying the Chrono-Color projection $\mathcal{P}(\cdot)$, the source dimension is reduced by a factor of T prior to JSCC encoding, enabling extreme compression while preserving task-critical temporal semantics.

\section{The Proposed ChronoSC Framework}
\subsection{Semantic Encoder: Chrono-Color Stacking}
\label{sec:chrono_stacking}
\begin{algorithm}[t]
\caption{Chrono-Color Stacking with Motion Extraction}
\label{alg:chrono_stacking}
\footnotesize
\begin{algorithmic}[1]
\Require Input Video $\mathbf{V} \in \mathbb{R}^{T \times 3 \times H \times W}$.

\State \textbf{Step 1: Background Modeling}
\State\quad $F_{\text{bg}} \gets \text{Mean}_t(\mathbf{V})$;
\State\quad $F_{\text{fg}} \gets |\mathbf{V} - F_{\text{bg}}|$; 

\State \textbf{Step 2: Motion Mask Generation}
\State $\mathbf{M} \gets \text{Max}_t(\text{Mean}_c(F_{\text{fg}})) > \tau$ \Comment{Thresholding with $\tau$}

\State \textbf{Step 3: Chrono-Color Encoding}
\For{$t \leftarrow 1$ to $T$}
    \State\quad $\theta_t \gets \frac{t}{T} \cdot \theta_{\max}$;
    \State\quad $F'_t \gets \text{HueRotate}(F_{\text{fg}}[t], \theta_t)$;
\EndFor

\State \quad$\mathbf{I}_{\text{sem}} \gets \max_{t}(F')$; 

\State \textbf{Output}: Stacked Image $\mathbf{I}_{\text{sem}}$, Motion mask $\mathbf{M}$.
\end{algorithmic}
\end{algorithm}

The core component of ChronoSC is the Chrono-Color Stacking semantic encoder, which compresses a video sequence into a single static image. Unlike traditional video codecs that rely on computationally expensive motion estimation (e.g., optical flow) or deep 3D convolutional networks, Chrono-Color Stacking employs a parameter-free, arithmetic-based projection that collapses the temporal dimension of length $T$ into the color space. This results in an extreme $T:1$ compression ratio while preserving task-critical temporal semantics through chromatic encoding. The procedure consists of three steps, summarized in Algorithm \ref{alg:chrono_stacking}.

\subsubsection{Adaptive background subtraction}
To isolate dynamic content from static redundancy, we first estimate the background frame as the temporal mean of the input video $\mathbf{V} = \{F_1, F_2, \dots, F_T\}$. The foreground motion features are then extracted via element-wise absolute difference:
\begin{equation}
    F_{\text{bg}} = \frac{1}{T} \sum_{t=1}^{T} F_t, \quad F_{\text{fg}, t} = |F_t - F_{\text{bg}}|.
\end{equation}
This operation acts as a lightweight motion-attention mechanism that suppresses static background regions while highlighting moving objects, which are typically most informative for VideoQA tasks. The use of simple arithmetic operations ensures low computational overhead, making the approach suitable for embedded and edge devices. We note that this background subtraction assumes a stationary camera, which is common in edge surveillance and monitoring scenarios. In cases involving camera motion (e.g., UAV platforms), global motion compensation could be applied as a preprocessing step.

\subsubsection{Temporal-to-Chromatic mapping (Hue Shift)}
To encode temporal order into the spatial domain, we introduce a differentiable Hue Shift operator $\mathcal{H}(\cdot)$. Each foreground frame is assigned a hue rotation angle
\begin{equation}
    \theta_t = \frac{t}{T} \cdot \theta_{\max}.
\end{equation}
The frame $F_{\text{fg}, t}$ is transformed from the RGB color space to the YIQ color space, rotated around the luminance (Y) axis to preserve brightness while shifting chromaticity, and then mapped back to RGB:
\begin{equation}
    F'_{t} = \mathcal{H}(F_{\text{fg}, t}, \theta_t).
\end{equation}


\subsubsection{Max Projection Aggregation}
Finally, to compress the sequence into a single semantic symbol $\mathbf{I}_{\text{sem}} \in \mathbb{R}^{3 \times H \times W}$, we apply a max projection operator along the temporal dimension. Specifically, for each channel $c$ and spatial location $(h, w)$, the value is computed as:
\begin{equation}
    \mathbf{I}_{\text{sem}}[c, h, w] = \max_{t \in \{1,\dots,T\}} \left( F'_t[c, h, w] \right).
\end{equation}
Unlike average pooling, which causes motion blur, this pixel-wise maximization preserves the high-frequency details and peak intensity of the moving object trails. The resulting $\mathbf{I}_{\text{sem}}$ is a standard RGB image that compactly encodes object trajectories, motion direction, and visual attributes, serving as a task-relevant semantic symbol for downstream transmission and inference.

\subsection{Robust Transmission: Motion-Aware Swin Transceive}
\label{sec:mast}
To reliably transmit the semantic image over noisy wireless channels, we propose MAST. Unlike conventional DeepJSCC schemes that allocate channel resources uniformly across the image, MAST explicitly prioritizes motion-relevant regions using the motion mask produced by Chrono-Color Stacking.

\subsubsection{The MAST architecture}
MAST extends SwinJSCC by incorporating a parallel motion guidance branch. The encoder consists of two streams:
\begin{itemize}
    \item \textbf{Content stream:} A Swin Transformer encoder extracts latent features $\mathbf{F}_{\text{content}} \in \mathbb{R}^{C \times \frac{H}{16} \times \frac{W}{16}}$.
    \item \textbf{Guidance stream:} A lightweight convolutional encoder processes the binary motion mask $\mathbf{M}$ to generate a spatial attention map $\mathbf{A}_{\text{motion}}$.
\end{itemize}

\subsubsection{Motion-aware feature modulation}
The motion mask is first downsampled to match the spatial resolution of the latent features and then projected into the channel dimension using a $1\times1$ convolution followed by a sigmoid activation:
\begin{equation}
    \mathbf{A}_{\text{motion}} = \sigma\Big(\text{Conv}_{1\times1}\big(\text{Downsample}(\mathbf{M})\big)\Big)
\end{equation}
yielding an attention map $\mathbf{A}_{\text{motion}} \in [0, 1]^{C \times \frac{H}{16} \times \frac{W}{16}}$.

To prioritize motion-relevant regions, the content features are modulated using a residual gating mechanism:
\begin{equation}
    \mathbf{F}_{mod} = \mathbf{F}_{\text{content}} \odot (\mathbf{A}_{\text{motion}} + \epsilon)
\end{equation}
where $\epsilon$ ensures a minimum bandwidth allocation for static regions, while regions with strong motion cues receive proportionally higher transmission capacity.

\subsubsection{Transmission and reconstruction}
The modulated latent features are power-normalized and transmitted over the wireless channel. At the receiver, a Swin Transformer–based decoder reconstructs the semantic image $\hat{\mathbf{I}}_{\text{sem}}$ from the noisy channel output. The transceiver is trained end-to-end using a motion-weighted reconstruction loss:
\begin{equation}
    \mathcal{L} = \mathcal{L}_{mse}(\mathbf{I}_{\text{sem}}, \hat{\mathbf{I}}_{\text{sem}}) \odot (\mathbf{1} + \alpha \mathbf{M})
\end{equation}
where the weighting factor $\alpha$ emphasizes reconstruction fidelity in dynamic regions that are critical for semantic reasoning.

\subsection{Semantic Decoder: Task-Oriented VQA}
\label{sec:semantic_decoder}
ChronoSC adopts a strictly task-oriented design in which the receiver directly performs reasoning on the reconstructed semantic image rather than attempting to recover the original video. To this end, we employ a pre-trained BLIP vision-language model as the semantic decoder.

BLIP consists of a Vision Transformer encoder that extracts visual features from the reconstructed image $\hat{\mathbf{I}}_{\text{sem}}$ and a text decoder that generates an answer given a question $Q$: 
\begin{equation}
    A = \text{BLIP}(\hat{\mathbf{I}}_{\text{sem}}, Q).
\end{equation}
To preserve modularity and enable reuse of foundation models, training is decoupled into two stages. The MAST transceiver is first trained to robustly transmit Chrono-Color images under varying channel conditions, after which BLIP is fine-tuned on clean Chrono-Color images to learn the mapping from chromatic motion cues to temporal semantics. During inference, the transmitter and receiver operate independently, and the fine-tuned BLIP model directly infers answers from noisy reconstructed semantic images. The full training and inference pipeline is summarized in Algorithm \ref{alg:training_process}.

\begin{algorithm}[t]
\footnotesize
\caption{Decoupled Training Strategy for ChronoSC}
\label{alg:training_process}
\begin{algorithmic}[1]
\Require Video dataset $\mathcal{D} = \{(\mathbf{V}_i, Q_i, A_i)\}_{i=1}^N$, pre-trained BLIP $\mathcal{M}_{\text{init}}$.
\Ensure Optimized MAST parameters $(\theta^*, \phi^*)$, and fine-tuned VLM $\psi^*$.

\Statex \textbf{Stage 1: Train MAST for Semantic Reconstruction}
\State \textit{Objective: Learn to transmit Chrono-Color images robustly.}
\For{epoch $= 1$ to $E_1$}
    \For{each batch $\mathbf{V}$ in $\mathcal{D}$}
        \State $(\mathbf{I}_{\text{sem}}, \mathbf{M}) \gets \mathcal{P}(\mathbf{V})$; \Comment{Generate Stacked Image}
        \State $\mathbf{Z} \gets f_\theta(\mathbf{I}_{\text{sem}}, \mathbf{M})$; \Comment{Motion-Aware Encoding}
        \State $\hat{\mathbf{Z}} \gets \text{Channel}(\mathbf{Z}, \text{SNR}_{\text{train}});$
        \State $\hat{\mathbf{I}}_{\text{sem}} \gets g_\phi(\hat{\mathbf{Z}})$; \Comment{Reconstruction}
        \State $\mathcal{L}_{\text{rec}} \gets \|\mathbf{I}_{\text{sem}} - \hat{\mathbf{I}}_{\text{sem}}\|^2 \odot (\mathbf{1} + \alpha \mathbf{M})$;
        \State Update MAST params $\theta, \phi \gets \text{Optimizer}(\mathcal{L}_{\text{rec}})$;
    \EndFor
\EndFor

\Statex \textbf{Stage 2: Train BLIP on Chrono-Color Domain }
\State \textit{Objective: Teach BLIP to understand "Time-to-Color" mapping.}
\State Initialize VLM $\mathcal{M}_\psi \gets \mathcal{M}_{\text{init}}$;
\For{epoch $= 1$ to $E_2$}
    \For{each batch $(\mathbf{V}, Q, A)$ in $\mathcal{D}$}
        \State $(\mathbf{I}_{\text{sem}}, \mathbf{M}) \gets \mathcal{P}(\mathbf{V})$ \Comment{Use clean Stacked Image}
        \State Predict answer: $\hat{A} \gets \mathcal{M}_\psi(\mathbf{I}_{\text{sem}}, Q)$;
        \State $\mathcal{L}_{\text{content}} \gets \text{CE}(\hat{A}, A)$;
        \State Update VLM params $\psi \gets \text{Optimizer}(\mathcal{L}_{\text{content}})$;
    \EndFor
\EndFor

\Statex \textbf{--- Inference Phase ---}
\State Transmit: $\mathbf{V} \xrightarrow{\mathcal{P}} (\mathbf{I}_{\text{sem}}, \mathbf{M}) \xrightarrow{\text{MAST}} \hat{\mathbf{I}}_{\text{sem}}$;
\State Reason: $\hat{A} = \mathcal{M}_{\psi^*}(\hat{\mathbf{I}}_{\text{sem}}, Q)$;

\State \textbf{Return} $\theta^*, \phi^*, \psi^*$.
\end{algorithmic}
\end{algorithm}

\section{Experiments}
\label{sec:experiments}

\subsection{Experimental Setup}
\label{subsec:setup}

\textbf{Dataset:} We evaluate ChronoSC on the CLEVRER dataset \cite{clevrer}, a diagnostic benchmark designed for video-based causal and physical reasoning. Our experiments focus on the Descriptive question subset, which assesses a model’s ability to perceive object attributes (e.g., color, shape, and material) as well as motion dynamics such as direction and velocity. All videos are processed at a resolution of $256 \times 256$.\\

\textbf{Implementation details and hyperparameters:}
The proposed framework is implemented in PyTorch and trained on a single NVIDIA RTX 3090 GPU. Training follows the decoupled strategy described in Section IV.
\begin{itemize}
    \item \textbf{1) MAST training (Stage 1):} Trained for 50 epochs (batch size 16, AdamW optimizer, initial lr $1 \times 10^{-4}$ with cosine annealing). We set $\alpha=0.5$ and latent dimension $C=32$.
  
    \item \textbf{BLIP fine-tuning (Stage 2):} We adopt the pre-trained BLIP-Base model (ViT-B/16 backbone) as the semantic decoder. BLIP is fine-tuned for 20 epochs on clean Chrono-Color stacked images using a learning rate of $2 \times 10^{-5}$.

\end{itemize}

\textbf{Baselines:} To assess the effectiveness of ChronoSC, we compare it against four representative transmission schemes:
\begin{enumerate}
    \item \textbf{H.264 + LDPC:} A conventional digital scheme where the video ($T=8$ frames) is compressed using H.264 and protected by LDPC codes ($1/2$ rate).
    \item \textbf{DeepJSCC (single frame):} Only the middle video frame is transmitted using a transformer-based DeepJSCC encoder \cite{swinjscc}, testing whether temporal information is necessary.

    \item \textbf{Averaged frame:} All frames are averaged into a single image before DeepJSCC transmission, serving as a simple temporal aggregation baseline.
    
    \item \textbf{Frame-by-Frame DeepJSCC:} Each of the $T=8$ frames is independently encoded using DeepJSCC and processed by Video-BLIP, representing a high-bandwidth oracle-style approach.
\end{enumerate}

\subsection{Results \& Discussion}
\label{subsec:results}
\begin{figure}[t]
    \centering
    \includegraphics[width=1\linewidth]{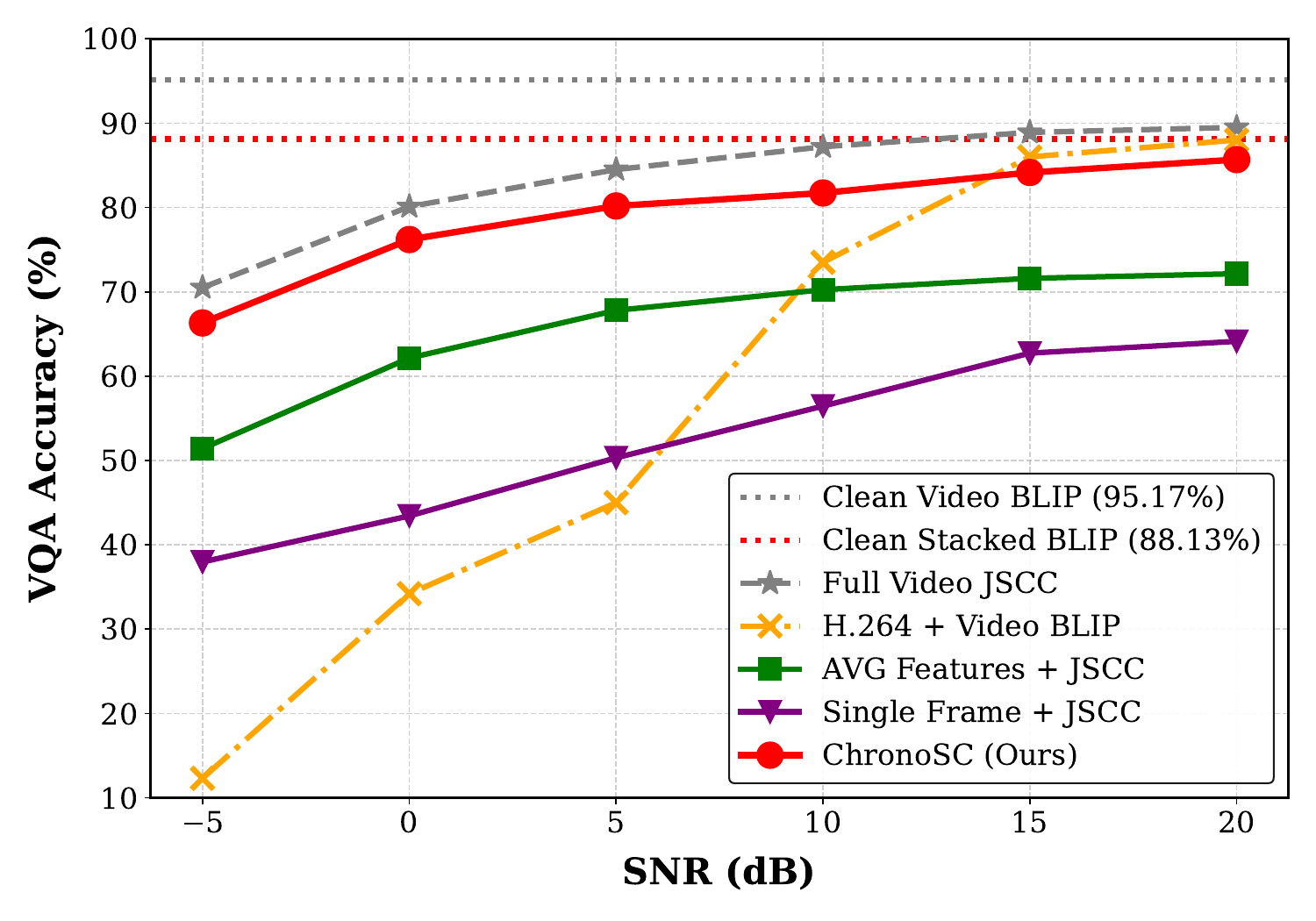}
    \caption{Comparison of VQA accuracy across different methods. }
    \label{fig:accuracy_comparison}
\end{figure}
\textbf{Impact of channel noise:} As shown in Fig. \ref{fig:accuracy_comparison}, ChronoSC excels in low-SNR regimes ($-5$ to $5$ dB). At $-5$ dB, conventional H.264+LDPC suffers a severe cliff effect, dropping to $12.3\%$ accuracy. In contrast, ChronoSC maintains $66.2\%$ via the graceful degradation of the MAST transceiver. Furthermore, ChronoSC outperforms the JSCC-based Averaged Frame baseline by $\sim 15\%$ across all SNRs, proving that Chrono-Color Stacking effectively preserves critical motion semantics (e.g., direction and velocity) lost during naive temporal aggregation.

\begin{table}[t]
\centering
\caption{\textbf{Performance Comparison at Extreme Noise (SNR = 0 dB).}}
\label{tab:main_comparison}
\resizebox{0.48\textwidth}{!}{%
\begin{tabular}{l c c c}
\toprule
\multirow{2}{*}{\textbf{Method}} & \multirow{2}{*}{\textbf{Input Data}} & \multirow{2}{*}{\textbf{BCR ($\downarrow$)}} & \textbf{Accuracy ($\uparrow$)} \\
 & & & {@ SNR = 0 dB} \\
\midrule
\multicolumn{4}{l}{\textit{\textbf{Baselines}}} \\
1. H.264 + LDPC & Video ($T=8$) & $\approx 1.6 \times 10^{-4}$ & $35.2\%$ \\
2. Single Frame & Image ($T=1$) & $\mathbf{1.6 \times 10^{-4}}$ & $43.4\%$ \\
3. Averaged Frame & Image ($T=1$) & $\mathbf{1.6 \times 10^{-4}}$ & $62.2\%$ \\
4. Full Video JSCC & Video ($T=8$) & $1.3 \times 10^{-3}$ & $\mathbf{80.1\%}$ \\
\midrule
\multicolumn{4}{l}{\textit{\textbf{Proposed Method}}} \\
\textbf{5. ChronoSC (Ours)} & \textbf{Stacked Img} & $\mathbf{1.6 \times 10^{-4}}$ & \underline{$76.2\%$} \\
\bottomrule
\end{tabular}%
}
\end{table}

\textit{Bandwidth efficiency and compression}: Table \ref{tab:main_comparison} reports performance under an extreme noise condition (SNR = $0$ dB), highlighting the trade-off between bandwidth and accuracy.
\begin{itemize}
    \item \textbf{Bandwidth Efficiency and Compression:} The Full Video JSCC oracle achieves the highest accuracy ($80.1\%$) but incurs a high bandwidth cost ($\text{BCR} \approx 1.3 \times 10^{-3}$). In contrast, ChronoSC attains a comparable accuracy of $76.2\%$ (only a $3.9\%$ drop) while achieving $\mathbf{8\times}$ less bandwidth ($\text{BCR} = 1.6 \times 10^{-4}$). Crucially, while this represents an $8\times$ reduction against the compressed baseline, it corresponds to a massive $\mathbf{192\times}$ bandwidth reduction compared to the transmission of \textbf{raw uncompressed video} ($T=8$ frames). This result confirms that Chrono-Color Stacking effectively compresses redundant temporal information into a compact spatial representation, making it highly suitable for bandwidth-starved edge environments where transmitting raw or standard-coded video is infeasible.
    
    
\end{itemize}

\subsection{Computational Complexity Analysis}
\label{subsec:complexity}

To explicitly validate the lightweight design of the proposed ChronoSC framework, we benchmarked the computational complexity of Chrono-Color Stacking against a standard 3D Convolutional Neural Network (C3D-like) backbone commonly used for spatiotemporal feature extraction. The benchmarking was conducted on a standard hardware setup using a 16-frame video input at a $128 \times 128$ resolution.

\begin{table}[h]
\centering
\caption{\textbf{Computational Complexity Evaluation}}
\label{tab:complexity}
\resizebox{0.48\textwidth}{!}{%
\begin{tabular}{l c c c}
\toprule
\textbf{Method} & \textbf{FLOPs} & \textbf{CPU Time (ms)} & \textbf{GPU Time (ms)} \\
\midrule
3D CNN (Baseline) & 24.9 G & 86.4 & 6.1 \\
\textbf{ChronoSC (Ours)} & \textbf{595.5 M} & \textbf{18.3} & \textbf{3.7} \\
\bottomrule
\end{tabular}%
}
\end{table}

As summarized in Table \ref{tab:complexity}, the Chrono-Color projection requires only $595.5$ MFLOPs and executes in $18.3$ ms on a CPU. This represents a dramatic $41.8\times$ reduction in computational complexity compared to the 24.9 GFLOPs demanded by the 3D CNN baseline. Crucially, ChronoSC operates $4.7\times$ faster on CPU hardware, directly addressing the scalability and energy constraints inherent to edge computing.

\begin{figure*}[t]
    \centering
    \definecolor{gtblue}{RGB}{0, 102, 204}
    \definecolor{badred}{RGB}{200, 0, 0}
    \setlength{\tabcolsep}{3pt} 

    \begin{minipage}{\linewidth}
        \centering
        \subfloat[\textbf{Input Video Frames} (Sampled at $t=1,3,5,7$)] {
            \includegraphics[height=2.75cm]{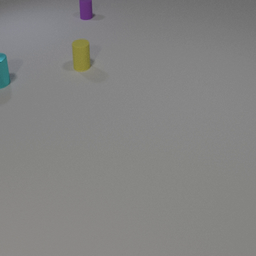}\hspace{0.25pt}
            \includegraphics[height=2.75cm]{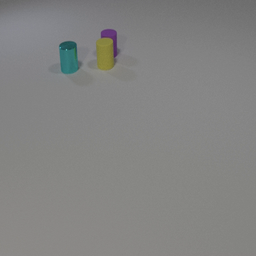}\hspace{0.25pt}
            \includegraphics[height=2.75cm]{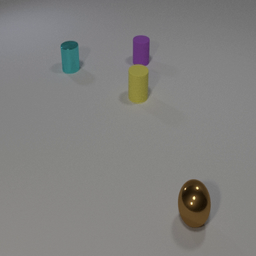}\hspace{0.25pt}
            \includegraphics[height=2.75cm]{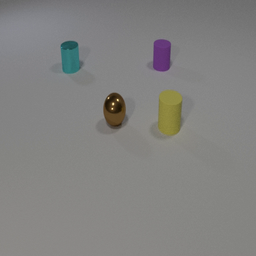}
            \label{subfig:frames}
        }
        \hfill \vrule width 0.05pt \hfill 
        \subfloat[\textbf{Chrono-Color Encoding} (Ours)] {
            \begin{minipage}[b]{0.15\linewidth}
                \centerline{\scriptsize (i) Original ($Tx$)}
                \centering
                \includegraphics[height=2.75cm]{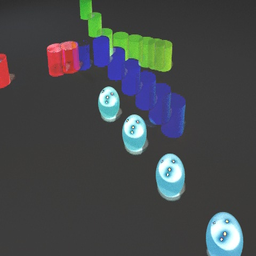}

            \end{minipage}
            \hspace{2pt}
            \begin{minipage}[b]{0.15\linewidth}
                \centerline{\scriptsize (ii) Rx @ \textbf{0dB}}
                \centering
                \includegraphics[height=2.75cm]{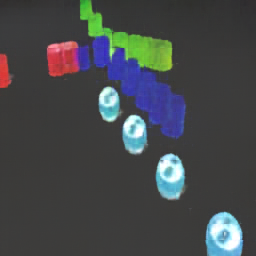}

            \end{minipage}
            \label{subfig:chrono}
        }
    \end{minipage}
    
    \vspace{0.1cm} 
    
    \begin{minipage}{\linewidth}
        \centering
        \small
        \renewcommand{\arraystretch}{1} 
        
        \resizebox{\linewidth}{!}{%
        \begin{tabular}{p{0.35\linewidth} c | c c c c | c}
            \toprule
            \multirow{2}{*}{\textbf{Question (Descriptive)}} & \textbf{Ground Truth} & \multicolumn{4}{c|}{\textbf{Baselines (@ SNR=0dB)}} & \textbf{Proposed} \\
            \cmidrule(lr){3-6} \cmidrule(lr){7-7}
             & (Label) & \shortstack{H.264\\+LDPC} & \shortstack{Single\\Frame} & \shortstack{Avg.\\Frame} & \shortstack{Full Video\\(Oracle)} & \textbf{ChronoSC} \\
            \midrule
            
            1. What is the color of the object colliding with the purple cylinder? & 
            \textcolor{gtblue}{\textbf{Yellow}} & 
            \textcolor{badred}{Grey \xmark} & 
            Yellow \cmark & 
            Yellow \cmark & 
            Yellow \cmark & 
            \textbf{Yellow} \cmark \\
            \midrule
            
            2. What is the shape of the \textbf{moving} metal object when the video ends? & 
            \textcolor{gtblue}{\textbf{Sphere}} & 
            \textcolor{badred}{Cube \xmark} & 
            \textcolor{badred}{Cylinder \xmark} & 
            \textcolor{badred}{Cylinder \xmark} & 
            Sphere \cmark & 
            \textbf{Sphere} \cmark \\ 
            \midrule
            
            3. What is the shape of the object that is \textbf{stationary} when the sphere \textbf{enters}? & 
            \textcolor{gtblue}{\textbf{Cylinder}} & 
            \textcolor{badred}{Sphere \xmark} & 
            \textcolor{badred}{Sphere \xmark} & 
            \textcolor{badred}{Cube \xmark} & 
            Cylinder \cmark & 
            \textbf{Cylinder} \cmark \\
            \bottomrule
        \end{tabular}%
        }
    \end{minipage}
    
    \caption{\textbf{Visualization of semantic transmission under extreme noise (SNR = 0 dB):} 
    (a) Sampled input video frames showing a moving sphere and a stationary cylinder.
(b) Chrono-Color Stacking encodes the temporal dynamics into a single static image (i). Despite severe channel noise (ii), the resulting color trails remain discernible, enabling the fine-tuned BLIP decoder to correctly infer motion and object attributes (bottom table).}
    \label{fig:qualitative_full}
\end{figure*}

Fig. \ref{fig:qualitative_full} provides qualitative insights into why ChronoSC succeeds under severe channel noise.

\begin{itemize}
    \item \textbf{Interpreting ``Rainbow Trails":} In Question 2 (“What is the shape of the moving metal object?”), the Single Frame baseline fails because a static snapshot cannot disambiguate moving and stationary objects. The Averaged Frame baseline introduces ghosting artifacts that blur object boundaries. In contrast, ChronoSC preserves a clear color-gradient trajectory corresponding to the moving sphere, enabling correct inference.
    \item \textbf{Adaptation of vision-language models:} The correct predictions produced by the fine-tuned BLIP model demonstrate that the VLM successfully learns the mapping from color gradients to motion semantics. This supports our hypothesis that pre-trained VLMs can interpret non-natural ''chrono-images” as a new visual language, effectively bridging low-level signal compression and high-level semantic reasoning.
\end{itemize}

A potential concern with Chrono-Color encoding is ambiguity between intrinsic object color and motion-induced hue shifts. However, qualitative results indicate that BLIP reliably distinguishes the two: intrinsic colors remain spatially consistent across object bodies, while motion cues manifest as directional color gradients along object boundaries. Fine-tuning enables the model to learn this distinction effectively.

\section{Conclusion}
\label{sec:conclusion}

We proposed ChronoSC, a task-oriented semantic communication framework that compresses temporal video dynamics into a single static image via Chrono-Color Stacking for efficient edge Video Question Answering. Experimental results show that ChronoSC achieves performance comparable to full-video transmission while reducing bandwidth usage by over two orders of magnitude and maintaining strong robustness under noisy channel conditions.

\section*{Acknowledgement}
This work was supported by VinUni's Student Research Grant Program AY24-25 CECS.2425.004.

\bibliographystyle{IEEEtran}
\bibliography{bibo}

@ARTICLE{SenuraTMC,
  author={Wanasekara, Senura Hansaja and Nguyen, Van-Dinh and Wong, Kok-Seng and Nguyen, M.-Duong and Chatzinotas, Symeon and Dobre, Octavia A.},
  journal={IEEE Trans. Mobi. Comput.}, 
  title={{SC-GIR}: Goal-oriented Semantic Communication via Invariant Representation Learning for Image Transmission}, 
  year={2025},
  volume={},
  number={},
  pages={1-15},
  doi={10.1109/TMC.2025.3600434}}

@article{cliff,
  author  = {Wiegand, Thomas and Sullivan, Gary J. and Bjontegaard, Gisle and Luthra, Ajay},
  title   = {Overview of the {H.264/AVC} video coding standard},
  journal = {IEEE Trans. Circ.  Syst.  Video Tech.},
  volume  = {13},
  number  = {7},
  pages   = {560--576},
  year    = {2003}
}

@article{semcom_survey,
  author  = {Qin, Zhijin and Tao, Xiaoming and Lu, Jianhua and Tong, Wen and Li, Geoffrey Ye},
  title   = {Semantic communications: Principles and challenges},
  journal = {arXiv preprint arXiv:2201.01389},
  year    = {2021}
}

@article{videoqasc,
  author  = {Li, Guangpeng and Wang, Si and Gao, Zhen and Guo, Qing and Li, Geoffrey Ye},
  title   = {{VideoQA-SC}: Adaptive Semantic Communication for Video Question Answering},
  journal = {IEEE J. Sel. Areas  Commun.},
  year    = {2025},
  note    = {(To appear)}
}

@article{swinjscc,
  author  = {Yang, Ke and Wang, Sixian and Dai, Jincheng and Qin, Xiaoqi and Niu, Kai and Zhang, Ping},
  title   = {{SwinJSCC}: Taming {Swin} Transformer for Deep Joint Source-Channel Coding},
  journal = {IEEE Trans. Cogn. Commun.  Networ.},
  year    = {2024},
  publisher={IEEE}
}

@inproceedings{blip,
  author    = {Li, Junnan and Li, Dongxu and Xiong, Caiming and Hoi, Steven},
  title     = {{BLIP}: Bootstrapping Language-Image Pre-training for Unified Vision-Language Understanding and Generation},
  booktitle = {Proc. Inter. Conf. Machine Learning (ICML)},
  year      = {2022}
}

@inproceedings{clevrer,
  author    = {Yi, Kexin and Gan, Chuang and Li, Yunzhu and Kohli, Pushmeet and Wu, Jiajun and Torralba, Antonio and Tenenbaum, Joshua B.},
  title     = {{CLEVRER}: Collision Events for Video Representation and Reasoning},
  booktitle = {Proc. Inter. Conf. Machine Learning (ICML)},
  year      = {2020}
}

@article{bourtsoulatze,
  author  = {Bourtsoulatze, Eirina and Kurka, David Burth and Gündüz, Deniz},
  title   = {Deep joint source-channel coding for wireless image transmission},
  journal = {IEEE Trans. Cogn. Commun.  Networ.},
  volume  = {5},
  number  = {3},
  pages   = {567--579},
  year    = {2019}
}

@article{mudeepsc,
  author  = {Xie, Huiqiang and Qin, Zhijin and Li, Geoffrey Ye},
  title   = {Task-oriented multi-user semantic communications for {VQA} task},
  journal = {IEEE Wire.s Commun. Lett.},
  volume  = {11},
  number  = {3},
  pages   = {553--557},
  year    = {2021}
}

@article{yosef,
  author  = {Yosef, Erez and Elmalem, Shay and Giryes, Raja},
  title   = {Video reconstruction from a single motion blurred image using learned dynamic phase coding},
  journal = {Scientific Reports},
  volume  = {13},
  pages   = {13625},
  year    = {2023},
  publisher={Nature Publishing Group}
}

@article{vdmmd,
  author  = {Zhong, Yixin and others},
  title   = {{VDM-MD}: Video Diffusion Model for Motion Deblurring},
  journal = {arXiv preprint arXiv:2501.12604},
  year    = {2025}
}

@article{wu2022slotformer,
  title={SlotFormer: Unsupervised Visual Dynamics Simulation with Object-Centric Models},
  author={Wu, Ziyi and Dvornik, Nikita and Greff, Klaus and Kipf, Thomas and Garg, Animesh},
  journal={arXiv preprint arXiv:2210.05861},
  year={2022}
}

@INPROCEEDINGS{le2020hierarchical,
  author={Le, Thao Minh and Le, Vuong and Venkatesh, Svetha and Tran, Truyen},
  booktitle={Proc. IEEE/CVF Conf. Comp. Visi. Patt. Recog. (CVPR)}, 
  title={Hierarchical Conditional Relation Networks for Video Question Answering}, 
  year={2020},
  volume={},
  number={},
  location = {Seattle, WA, USA, 2020},
  pages={9969-9978},
  doi={10.1109/CVPR42600.2020.00999}}
\end{document}